\documentclass{article}

\usepackage[dvipsnames,table]{xcolor}
\usepackage{import}
\usepackage{arxiv}
\usepackage{multirow}
\usepackage{siunitx} 
\usepackage{hyperref}       
\usepackage{url}            
\usepackage{booktabs}       
\usepackage{amsfonts}       
\usepackage{nicefrac}       
\usepackage{microtype}      
\usepackage{mathtools}
\usepackage{bm} 
\usepackage[numbers]{natbib}
\usepackage{amsmath}
\usepackage{graphicx}
\usepackage{dsfont}
\usepackage{titlesec}
\usepackage{colortbl}
\usepackage{tabularx}
\usepackage{dblfloatfix}
\usepackage[capitalise]{cleveref}
\newcommand{\codelink}{https://github.com/fy0cc/factor-inference-net/}
\renewcommand\cite{\citep}
\definecolor{mygrey}{rgb}{0.1,0.1,0.1}
\definecolor{rev}{rgb}{0,0,0}

\DeclarePairedDelimiter\floor{\lfloor}{\rfloor}

\newcommand{\be}{\begin{equation}}
\newcommand{\ee}{\end{equation}}
\newcommand{\ba}{\begin{align}}
\newcommand{\ea}{\end{align}}
\newcommand{\bea}{\begin{eqnarray}}
\newcommand{\eea}{\end{eqnarray}}

\usepackage{amsmath}

\newcommand{\va}{{\mathbf{a}}}

\newcommand{\ve}{{\mathbf{e}}}

\newcommand{\vh}{{\mathbf{h}}}

\newcommand{\vm}{{\mathbf{m}}}

\newcommand{\vs}{{\mathbf{s}}}

\newcommand{\vx}{{\mathbf{x}}}

\newcommand{\vF}{{\mathbf{F}}}

\newcommand{\vT}{{\mathbf{T}}}

\newcommand{\vW}{{\mathbf{W}}}

\title{Generalization of graph network inferences in higher-order graphical models}

\author{ 
Yicheng Fei\\
Rice University\\
Houston, TX 77005 \\
\text{yf17@rice.edu} \\
\And
Xaq Pitkow\\
Baylor College of Medicine\\
Rice University\\
Center for Neuroscience and Artificial Intelligence\\
Houston, TX 77030 \\
\text{xaq@rice.edu} \\
}

\begin{document}
\maketitle

\begin{abstract}
Probabilistic graphical models provide a powerful tool to describe complex statistical structure, with many real-world applications in science and engineering from controlling robotic arms to understanding neuronal computations. A major challenge for these graphical models is that inferences such as marginalization are intractable for general graphs. These inferences are often approximated by a distributed message-passing algorithm such as Belief Propagation, which does not always perform well on graphs with cycles, nor can it always be easily specified for complex continuous probability distributions. Such difficulties arise frequently in expressive graphical models that include intractable higher-order interactions. In this paper we define the Recurrent Factor Graph Neural Network (RF-GNN) to achieve fast approximate inference on graphical models that involve many-variable interactions. Experimental results on several families of graphical models demonstrate the out-of-distribution generalization capability of our method to different sized graphs, and indicate the domain in which our method outperforms Belief Propagation (BP). Moreover, we test the RF-GNN on a real-world Low-Density Parity-Check dataset as a benchmark along with other baseline models including BP variants and other GNN methods. Overall we find that RF-GNNs outperform other methods under high noise levels.
\end{abstract}


\section{Introduction}\label{sec:intro}
To draw conclusions about individual variables of interest in a task, we must marginalize out all other unobserved variables. Such exact inference computations are often infeasible in high-dimensional latent spaces, due to their exponential complexity. Conveniently, the latent space to be marginalized is often decomposable due to conditional dependencies between variables, a structure that can be described by a probabilistic graphical model (PGM) \cite{koller2009probabilistic}.

This decomposable structure may allow us to perform difficult global calculations using simpler computations on subsets of variables. {\color{rev}This makes probabilistic graphical models a compelling framework for describing both machine and biological intelligence. Early graphical models used pairwise interactions, as in Hopfield networks \cite{hopfield1982neural} and Boltzmann machines \cite{sherrington1975solvable}, to model inference, learning, and memory. 
However, basic pairwise interaction models may not be compact descriptions of complex data patterns seen in real data, whether in machine learning \cite{ranzato2010modeling,hinton2010learning}, neuroscience \cite{beggs2003neuronal,Ganmor.Schneidman.2011,shimazaki2015simultaneous}, biochemical networks \cite{ritz2014signal}, or social networks \cite{centola2018tipping,milojevic2014team,iocopini2019simplicial}.
By including higher-order interactions we increase model flexibility, but the number of possible interactions grows combinatorially with the interaction order, which contributes to the difficulty of exact inference in such models. This difficulty can be reduced by reasonable prior assumptions about locality and sparsity of natural interactions, resulting in a sparse higher-order graphical model.} 

{\color{rev}
Message-passing algorithms take advantage of this sparse graph structure to simplify computations.} Such approaches are used by algorithms like Belief Propagation (BP) \cite{pearl1988probabilistic} and Expectation Propagation (EP) \cite{minka2001expectation}, which are widely used approaches to computing or approximating marginal probabilities using distributed computation. BP is guaranteed to yield exact results if the graph has a tree structure. However, on general graphs with loops, which are likely to be better descriptors of real data, these algorithms can make substantial approximation errors or even fail to converge.

Unfortunately, higher-order factor graph may have more loops than pairwise graphs with as many interactions, and since standard local message-passing algorithms like BP suffer in the presence of loops they are likely to perform worse on many real-world graphical models. Even when applied to higher-order trees, message updates for BP usually don't have closed-form solutions, so running exact BP on these graphs becomes impractical. 

To mitigate these drawbacks of algorithms like BP and to provide an alternative on loopy graphs without analytical update formulas, in this work we present the Recurrent Factor Graph Neural Network (RF-GNN), a flexible recursive message-passing algorithm for fast approximate inference. Our method applies to a large family of higher-order graphical models with a wide range of graph structures and parameter values, including very loopy ones. In the spirit of message-passing algorithms like BP, we use a Graph Neural Network (GNN) \cite{scarselli2009graph,li2015gated} to perform message-passing on factor graphs iteratively. We train this network to compute sufficient statistics of all univariate marginal probabilities simultaneously for each instance of a distribution generated from a parametrized family of PGMs.

To study the performance of RF-GNNs, we run numerical experiments on two artificial datasets where we can calculate ground truth: Gaussian Graphical Models (GGM) and a small binary spin-glass system with third-order interactions. GGMs only have pairwise interactions, but the ground truth marginals can easily be computed without message-passing for very large graphs. 
In addition, we construct a dataset of more complex continuous PGMs with pairwise and third-order interactions. Since closed-form marginals do not exist for this interesting model class, we train an RF-GNN to predict univariate statistics estimated by Markov Chain Monte Carlo (MCMC) sampling. We also test the performance of the RF-GNN on a Low-Density Parity-Check (LDPC) decoding dataset with a very loopy factor graph as a real-world example and compare with previous methods.

Our experiments show that a trained RF-GNN has better performance than BP on an in-distribution test dataset, even when restricting the comparison to only those graphs for which BP dynamics converge. 
We also show that our model generalizes reliably out of distribution to probabilistic graphical models of different sizes than the training set.
By looking at how the error distribution depends on two graph metrics---average shortest path length and cluster coefficient---we find that an RF-GNN outperforms BP particularly well on graphs with small average shortest path length and large cluster coefficients, which are common properties of many real world graphs \cite{watts1998collective}. 
This suggests there is potential for using an RF-GNN as an approximation inference method on real-world PGMs with higher-order interactions. 

{\color{rev}In Section 2 we provide key background material about probabilistic graphical models and Graph Neural Networks. In Section 3 we first define our Recurrent Factor Graph Neural Network as a message-passing algorithm to calculate approximate marginal distribution in a family of graphical models. Then we describe four types of graphical models with different attributes and complexity on which we tested our model. In Section 4 we show the performance of RF-GNNs on these four datasets and compare with Belief Propagation. Of particular interest, we analyze how RF-GNNs generalize to larger graphs and how different graph attributes affect generalization performance. 
In Section 5 we discuss related message-passing inference algorithms for loopy graphs, and discuss other types of GNNs that apply to graphs with higher-order structure. Finally, in Section 6 we discuss potential limitations and extensions of our framework.}

\section{Background}\label{sec:background}
\subsection{Probabilistic graphical models}
Probabilistic graphical models describe multivariate probability distributions using graphs in which nodes represent variables. Nodes are connected if the corresponding variables are statistically dependent when conditioning on all other variables. 
{\color{rev} Different types of graphs represent different factorization structures or conditional independence relationships within the joint probability density.} Examples include Bayesian networks with directed acyclic graphs, Markov random fields using undirected graphs, and factor graphs which are undirected bipartite graphs connecting variable nodes and factor (interaction) nodes. 

\begin{figure}
    \centering
    \includegraphics[width=0.25\linewidth]{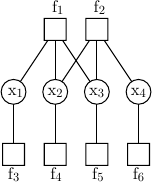}
    \caption{Diagram of a factor graph with 4 variable nodes and 6 factor nodes. Factors 1 and 2 are three-way factors each involving three variables. Factors 3 to 6 are singleton factors each only involving one variable.}
    \label{fig:factorgraph-diagram}
\end{figure}

Here we concentrate on the latter type, the factor graph \cite{frey1997factor}, which expresses a joint probability density as a product of local factors, $p(\vx)\propto \prod_{\alpha} f_{\alpha}(\vx_{\alpha})$, each involving only a subset of variables $\vx_\alpha$.
These models introduce a second type of node in the graph, the factor node, one for each local factor $f_{\alpha}$. Each factor node is only connected to the variable nodes involved in the interactions. An example of such a factor graph is depicted in \cref{fig:factorgraph-diagram}. 

{\color{rev}In this work we consider probabilistic graphical models within the exponential family \cite{pitman1936sufficient,Koopman.1936.sufficient}, a broad class of probability distributions widely used in statistical analysis. The density of such a distribution can be parameterized by a vector of natural parameters $\boldsymbol{\phi}$, 
\begin{equation}
    p(\vx) = \frac{1}{Z(\boldsymbol{\phi})}\exp(\boldsymbol{\phi}^\top \vT(\vx)),
    \label{eq:exponentialFamily}
\end{equation}
where $\vT(\vx)$ is the vector of sufficient statistics and $Z(\boldsymbol{\phi})$ is a normalization constant. In general, any single component of $\vT(\vx)$ can be an arbitrary function of all $\vx$. 
In a factor graph, however, these functions are restricted, with each involving only a subset of variables, $\vx_\alpha$, that are connected through a factor node. Each individual factor can then be parametrized with its own vector of natural parameters $\boldsymbol{\phi}_\alpha$:
\begin{equation}\label{eq:pgm-factor-graph}
    p(\vx) = \frac{1}{Z(\boldsymbol{\phi})} \prod_{\alpha} \exp(\boldsymbol{\phi}_{\alpha}^\top \vT_{\alpha}(\vx_{\alpha})).
\end{equation}
A PGM is then represented by sets of factor nodes, variable nodes, edges, and natural parameters as features on factor nodes, $\{\mathcal{F},\mathcal{V},\mathcal{E}, \{\boldsymbol{\phi}_{\alpha}\}\}$. 
}

{\color{rev}
Another, more familiar, representation of a PGM is in terms of expectations of the statistics $\mathbb{E}_{p(\vx)}[\vT(\vx)]$. These expectations are called expectation parameters, such as means and variances for a Gaussian. When combined with a maximum entropy principle, the expectation parameters implicitly fully specify the probability distribution. Expectation parameters and natural parameters are closely related to each other: the natural parameters are the Lagrange multipliers that enforce those expectations.
}

{\color{rev}
Relating these parameters is hard, but often important: the process of inference can be viewed, essentially, as converting between natural parameters and expectation parameters \cite{wainwright2008graphical,koller2009probabilistic}. Marginalization is one example of computing expectations of a subset of variables given natural parameters for the full PGM. The difficulty of this computation is what motivated our efforts to approximate it by a graph network.}

{\color{rev}
Message-passing algorithms like belief propagation exploit the conditional independence between variables implied by the graph structure so that the multidimensional integration could be performed separately in orthogonal subspaces. In this way, the computation complexity only depends on the maximum node degree and is independent of graph size. 
However, BP updates assume all neighbors are independent, which is not true in loopy graphs. \cite{yedidia2000generalized} shows that BP only converges to fixed points of the Bethe free energy, which is an approximation of the Gibbs free energy that only involves pairwise interactions. By generalizing the Bethe free energy to Kikuchi free energy which involves more complex clusters of nodes and corresponding messages between clusters, \cite{yedidia2000generalized} defines a generalized BP (GBP) algorithm that achieves better convergence and more accurate marginals on graphs with tight loops like lattices.
Our method currently works on regular factor graphs, but could be extended to more complicated Kikuchi clustering approximations.
}

%

Exact inference in tree graphs can be performed by iteratively marginalizing out the leaves of the tree and propagating this information along the graph. This iterative algorithm is called belief propagation (BP), and has been applied with some success even on graphs with loops \cite{pearl1988probabilistic,bishop:2006:PRML}. Message updates in BP include multivariate integration, and don't always have closed-form solutions. (Fully factorized) Expectation Propagation (EP) \cite{minka2001expectation} is an approximation to mitigate this issue by projecting the outgoing message to some convenient user-chosen parametric family.

\subsection{Graph Neural Networks}
Graph Neural Networks (GNN) are artificial neural networks implementing a message-passing operation on a graph \cite{scarselli2009graph}. A GNN updates each node's representation based on aggregated messages from its neighbors.
Each node $i$ represents information as a vector $\vh_i^t$ that evolves over time (or layer) $t$, and edges are assigned a vector weight $\ve_{ij}$. The updated representation at time or layer $t+1$ can be described by: 
\begin{equation}
    \vspace{-2mm}
    \vh_i^{t+1} = \mathcal{U}\Big(\vh_i^t\,, \!\!\!\bigsqcup_{j\in N(i)\backslash\{j\}}\!\!\!\!\!\! \mathcal{M}(\vh_j^t, \vh_i^t, \ve_{ij})\Big)
    \label{eq:GNNdynamics}
\end{equation}
where every message $\mathcal{M}(\vh_j^t, \vh_i^t, \ve_{ij})$ from neighbor $j$ to node $i$ along edge $ij$ is first calculated using a common trainable nonlinear message function $\mathcal{M}$, then messages from all neighbors are combined by a permutation-invariant aggregation function $\bigsqcup$ (e.g. summation), before being used to update each target node through another trainable update function $\mathcal{U}$. 
{\color{rev}Update functions based on a Gated Recurrent Unit (GRU) \cite{cho2014learning} or Long Short Term Memory \cite{hochreiter1997long} provide a long term memory for each node state.
Note that in other applications, the superscript $t$ can index a feedforward layer in a stacked GNN, where different layers have different weights for the message and update function $\mathcal{M}$ and $\mathcal{U}$ 
\cite{Velickovic.Bengio.2017,kipf2016semi}. Here, however, we use a recurrent GNN implementation where $t$ represents time, although this is equivalent to tying these layers' weights as in an unrolled Recurrent Neural Network (RNN) with message and update functions and factor parameters that are shared across layers \cite{li2015gated,gilmer2017neural}. 
}
{\color{rev}After passing messages throughout the graph for a certain number of time steps, a global or local readout network is used to extract the information we need, depending on the task.}
We recommend \cite{Zhou.Sun.2020} for a review of methods and applications of GNNs.

\section{Methods}\label{sec:methods}
\subsection{Recurrent Factor Graph Neural Network}
{\color{rev}To make more accurate inferences on probabilistic models with loopy factor graphs, we present the Recurrent Factor Graph Neural Network (RF-GNN), a network that takes in a factor graph representation of a probabilistic model from an exponential family and predicts approximate single-variate marginal distributions of every variable.
The RF-GNN learns a message-passing algorithm on the factor graph in which message functions and updates are executed by neural networks rather than multivariate integration. When run, this message-passing alters its node states iteratively as a dynamical system, operating essentially as a structured RNN, that upon convergence has node states that encode approximate marginals. As a flexible message-passing algorithm, once trained, an RF-GNN can be applied to graphical models in the same parametric family with various sizes and structures.
}

\begin{figure}
    \centering
    \includegraphics[width=\linewidth]{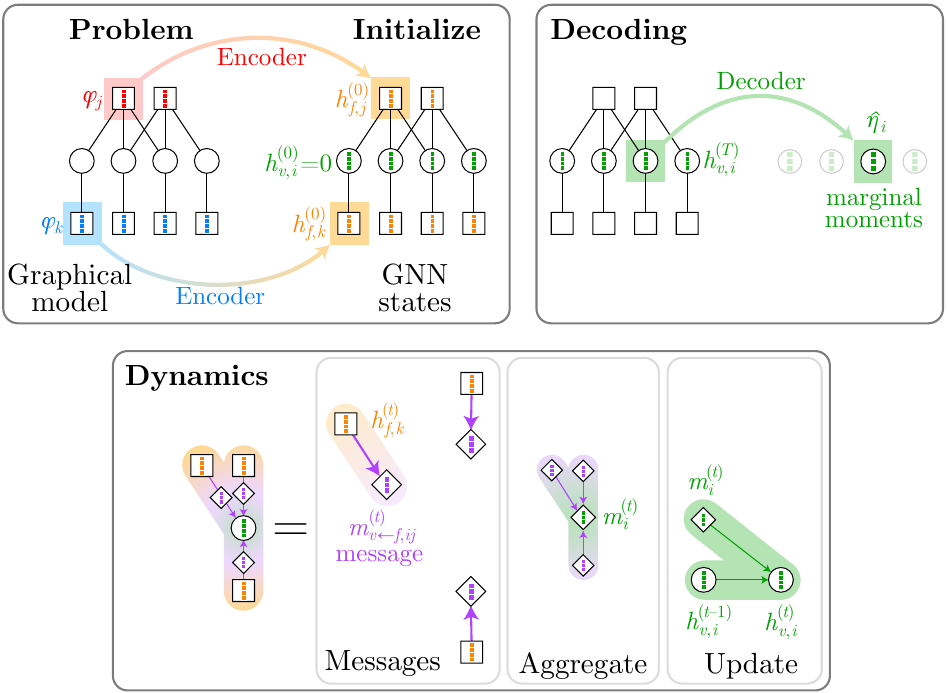}
    \caption{Diagram of an RF-GNN. Firstly, the variable and factor node features of the input PGM $\phi_j, \phi_k$s are encoded into latent features $h_{f,j}^{(0)},h_{f,k}^{(0)}$s using two separate encoders. Then a bidirectional message-passing network specified by the message and update neural networks recurrently update node representations. Finally, we decode the desired features of all single-variate marginal distributions $\hat{\eta}_i$ with a common decoder network}
    \label{fig:diagram}
\end{figure}

{\color{rev}Using the diagram in \cref{fig:diagram}, we now explain the core overall structure of an RF-GNN. After this overview, we will then describe the details of our network that we chose to create a concrete implementation of an RF-GNN.} Firstly, in order to define this class of message-passing algorithms on factor graphs, we define a GNN on generic bipartite graphs with two distinct types of nodes, unlike a typical GNN that treats all nodes equally \cite{kipf2016semi,gori2005new,li2015gated}. Our two node types are for variables and factors, represented by hidden vectors $\vh_{v,i}^{(t)}$ and $\vh_{f,i}^{(t)}$, with subscripts $v$ and $f$  for variable nodes and factor nodes. {\color{rev} At $t=0$, every node's latent state is initialized, and then the variable and factor node states $\vh_{v,i}^{(t+1)}, \vh_{f,i}^{(t+1)}$ are updated recursively from the values of the previous time step through message-passing. The message-passing alternates between updating variable nodes and factor nodes using their respective message and update functions. Each stage follows standard GNN dynamics, as shown in \cref{fig:diagram} (Dynamics) and described above (Eq~\ref{eq:GNNdynamics}).
After $T$ time steps determined by a convergence criterion, we use a common decoder network to decode the target from each variable $\vh_{v,i}^{(T)}$ simultaneously. In this work, we train our decoder network to predict the marginal moments of our training data from node state $\vh_{v,i}^{(T)}$. 
}


\subsection{Architectural details for an RF-GNN}

{\color{rev}Given this general algorithmic framework, we will now explain our design choices for the algorithm's components. In principle, our autonomous dynamical system could use either continuous time, like Neural Ordinary Differential Equations (neural ODEs) \cite{chen2018neural}, or discrete time, like many RNNs. We opt for a discrete-time RNN for computationally efficiency.
We choose GRUs \cite{cho2014learning} for the update operations as they have proven to be expressive enough while remaining more computationally efficient than LSTMs \cite{hochreiter1997long}. We add LayerNorm \cite{Ba.Hinton.2016} to the GRU state update, since that helps regularize the range of values in intermediate timesteps and avoid the common problem that activations tend to grow or shrink in a recurrent system.}

{\color{rev}We now describe the mathematical details of an RF-GNN, following the flow of computation. The inputs of an RF-GNN are graphical models $\{\mathcal{F},\mathcal{V},\mathcal{E}, \{\boldsymbol{\phi}_{\alpha}\}\}$ within the exponential family, parametrized by their natural parameters $\{\boldsymbol{\phi}_{\alpha}\}$, as described in \cref{eq:pgm-factor-graph}.
}


{\color{rev}
There are two ways of incorporating local potentials into the latent representations. One way is to start from a common initialization independent of the potentials, and provide the potential's parameters as additional inputs for every step in the message and update functions, just like belief propagation. This initialization is simpler in one way, but then we need more complex message functions and update functions that take the local potentials as inputs. 
Another way is to embed the potentials in the {\it initial states} of all factor nodes, and let the dynamics run autonomously from this initial condition. The potentials then only affect the inference by how the initial state shapes the dynamics. This actually corresponds to an interpretation of belief propagation as iterative reparameterization \cite{wainwright2003tree}. Here we choose a combination of both methods, as described in \cref{eq:initialization,eq:feature-matrix,eq:activation}.
The representation of each factor node is divided into two parts: a dynamic latent state $\vh_{f,\alpha}^{(t)}$ and a static feature matrix $\vF_{\alpha}$. The first is recurrently updated through message-passing, while the second serves as a constant input in the update function to transform the aggregated messages \cref{eq:activation}. We use two separate multilayer perceptron (MLP) \cite{rosenblatt1958perceptron} encoders to map the natural parameters into the initial state $\vh_{f,\alpha}^{(0)}$ and the feature matrix $\vF_{\alpha}$. Different network weights are used for different factor types.}
\begin{align}
    \vspace{-2mm}
    \vh_{f,\alpha}^{(0)} = \text{Encoder1}(\boldsymbol{\phi}_{\alpha}) \label{eq:factor-init}\\ 
    \vF_{\alpha} = \text{Encoder2}(\boldsymbol{\phi}_{\alpha}) \label{eq:feature-matrix}
\end{align}
The hidden states for variable nodes only have a recurrently updated part, which we initialize with zero vectors:
\begin{align}
    \label{eq:initialization}
    \vh_{v,j}^{(0)} = \mathbf{0}
\end{align}


At each iteration $t$, we calculate the message from variable $j$ to factor $\alpha$ as a linear projection of $\vh_{v,j}^{(t-1)}$, which is then averaged with messages from all neighbors $j \in N(\alpha)$ to form a summary message. The summary message is then transformed by a factor feature matrix $\vF_{\alpha}$ in a similar way as \cite{gilmer2017neural} to produce the activation $\va_{\alpha}^{(t)}$. We input $\va_{\alpha}^{(t)}$ along with the old state $\vh_{f,\alpha}^{(t-1)}$ into the GRU unit at factor node $\alpha$ to calculate the new latent state $\vh_{f,\alpha}^{(t)}$.
\begin{align}
    \vspace{-2mm}
    &\vm_{f,j\rightarrow \alpha}^{(t)} = \vW_{f,\alpha} \vh_{v,j}^{(t-1)} &\text{message} \label{eq:message} \\ 
    \addlinespace
    &\va_{f,\alpha}^{(t)} = \vF_{\alpha} \cdot \frac{1}{\|N(\alpha)\|}\sum_{j\in N(\alpha)} \vm_{f,j\rightarrow \alpha}^{(t)} &\text{aggregated message} \label{eq:activation}\\
    &\vh_{f,\alpha}^{(t)} = \text{GRU}_f(\vh_{f,\alpha}^{(t-1)},\va_{f,\alpha}^{(t)} ) &\text{state update} \label{eq:gru}
\end{align}
The updates for variable nodes are similar except that \Cref{eq:activation} is replaced by $\va_{v,j}^{(t)} = \frac{1}{\|N(j)\|}\sum_{\alpha\in N(j)} \vm_{v,\alpha \rightarrow j}^{(t)}$ since there is no feature associated with variables. 

We also explored the use of more complicated message functions like a two-layer MLP instead of linear transformation on $\vh_{v,j}^{(t)}$ to replace \cref{eq:message} and flexible message-passing mechanisms like Graph Attention Networks \cite{Velickovic.Bengio.2017,brody2021attentive}. However, they do not show significant performance gain on any of the datasets described in \Cref{sec:dataset}, so we retain the simple linear message function.

The latent states of factor and variable nodes $\vh_{f}^{(t)}, \vh_{v}^{(t)}$ are updated iteratively until step $T$ following \cref{eq:message,eq:activation,eq:gru}.
As discussed earlier, we would like our model to learn an iterative algorithm that converges to the target, however many steps it takes. So instead of choosing the number of time steps $T$ to be fixed, which we found often yielded RNN dynamics that passed through the target output without stopping, we randomly sample a readout time $T$ from a range, so the network cannot rely on any particular readout time. At that point, we use a decoder MLP to read out simple statistics of the univariate marginal distributions for each variable as our target.
\begin{equation}
    \boldsymbol{\eta}_j = \text{Decoder}(\vh_{v,j}^{(N)}).
\end{equation}
{\color{rev}For example, in Gaussian Graphical Models, we use the decoder to predict the inverse variance of the marginal distributions for each variable in the graph simultaneously. In continuous third-order graphical models, we use the first four central moments, which are expectation parameters, as our target.
We target the expectation parameters or other simple statistics because they are a convenient way of capturing information about the local marginals, and they are easy to extract from samples from the training distributions.
}
\subsection{Datasets}\label{sec:dataset}
For tractability, many probabilistic models are based on pairwise interactions. Other PGMs like Bayesian Networks can capture higher-order interactions directly through dependence on multiple parents, but such models are often decomposed into additive pairwise interactions. Pairwise models would require many nonlinear auxiliary hidden variables to capture real-world data complexity such as multiplicative lighting, perspective transforms, triple synapses, or especially gating. 
In contrast, higher-order interactions capture some of these interactions directly. For example, third-order multiplicative interactions can be effectively viewed as soft gating operations, a crucial, common operation in machine learning, as seen in LSTMs \cite{hochreiter1997long}, GRUs \cite{cho2014learning}, and transformer networks \cite{vaswani2017attention}. These third-order models provide more modeling power while remaining more interpretable statistically. Thus here we choose continuous graphical models with third-order interactions as a distribution family of high interest. Such distributions fall into the category of applications where traditional inference algorithms are infeasible but our GNNs still apply.

{\color{rev}We construct four PGM datasets (first four lines in \Cref{tab:dataset}) with increasing complexity and report the performance of the RF-GNN on these datasets compared to BP. Each dataset consists of a training set and several test sets. The training set includes graphical models with a fixed number of variables, but with diverse structures and factor parameters. Each test set is constructed in the same way as the training set, but with a different graph size.}


{\color{rev}First, we construct three datasets of PGMs with known ground truth marginal distributions and closed-form BP update formulas: Gaussian Graphical Model (GGM) with tree structure, Gaussian Graphical Model with arbitrary structure,} and binary Third-order Graphical Models. These datasets allow us to compare the performance of our model with Belief Propagation. {\color{rev} Notably, since BP produces exact marginals on tree graphs, performance on the GGM-tree dataset serves a proof of concept that the RF-GNN is highly accurate on tree graphs, even when tested on larger graphs not seen during training.} Second, we build a dataset of continuous PGMs with third-order interactions to test our model on a more complicated and highly interesting class of third-order PGMs.  {\color{rev}Finally, as a benchmark to compare the RF-GNN to other GNN-based models for PGMs, we test the RF-GNN on a Low-Density Parity-Check (LDPC) decoding dataset consisting of binary PGMs with a fixed structure but different parameters. Each of these datasets is described in greater detail below.
}

\subsubsection{Testing generalization on new graphs}
{\color{rev}
Flexible recurrent message-passing algorithms have the property of that we can apply them to graphs of sizes not seen during training. Here, we define how we measure an RF-GNN's ability to generalize to different sized graphs.
Every PGM dataset consists of a training set and several test sets. The training data $\mathcal{D}_{\rm train}^N=\{(\mathcal{G}_i,\boldsymbol{\eta}_i) \| \mathcal{G}_i \sim \mathcal{P}^G(N), \boldsymbol{\eta}_i \sim \mathcal{P}^{\eta}(N) \}$ is composed of PGMs with the same size $N$, but with a different graph-generating process $\mathcal{P}^G(N)$ and parameter-generating process $\mathcal{P}^{\eta}(N)$. We use the same processes $\mathcal{P}^G(M)$ and $\mathcal{P}^{\eta}(M)$ to generate test set $\mathcal{D}_{\rm test}^M$ of size $M$. We define the generalization capability of the RF-GNN to a different, especially larger, size $M$ as the performance of an RF-GNN trained on $\mathcal{D}_{\rm train}^N$ and evaluated on $\mathcal{D}_{\rm test}^M$. 
}

\newcommand{\scalelabel}[1]{\multicolumn{1}{l}{[$\times 10^{#1}$]}}
\sisetup{detect-weight=true,detect-inline-weight=math}

\newcommand{\longcell}[2][l]{%
  \begin{tabular}[#1]{@{}l@{}}#2\end{tabular}}
\rowcolors{1}{white}{mygrey!12}

\begin{table}[h]
\begin{center}
\caption[]{Properties of datasets. The first column describes the type of variables (discrete or continuous). {\color{rev}The second column indicates the computational complexity needed to obtain marginals for training.} The third column describes if BP performs exact or approximate inference, and the forth column indicates whether BP's message-passing (MP) update has a closed-form formula. The last column denotes the highest factor order of PGMs in the dataset. }\label{tab:dataset}
    \begin{tabular*}{\textwidth}{r|llllc}
        \toprule%
        \textbf{Dataset} & \textbf{Variable} & \textbf{\longcell{Complexity of\\marginals}} & \textbf{\longcell{BP \\inference}}& \textbf{MP form} & \textbf{\longcell{Max\\Order\ \ \ }} \\
        \midrule
        GGM-tree & continuous& linear & exact & closed-form & 2 \\
        GGM-all& continuous& cubic or better & approx.& closed-form& 2 \\
        3-binary & binary & exp. & approx. & closed-form & 3\\
        3-continuous & continuous& \longcell{exp. and approx. }& approx. & approx. & 3 \\
        LDPC & binary & not needed & approx. & closed-form & 6 \\
        \bottomrule
    \end{tabular*}
\end{center}
\end{table}


\subsubsection{Gaussian Graphical Model}\label{dataset:ggm}
As toy examples, we generate two datasets of random Gaussian Graphical Models (GGM): {\color{rev}\textit{GGM-tree}, and \textit{GGM-all}. The first dataset consist only tree graphs as a proof of concept. The second dataset includes graphs with diverse graph structures.} Since both exact marginalization and BP have closed-form solutions on GGMs, they serve as a convenient test for comparing an RF-GNN with BP, and allow us to test the generalization performance of an RF-GNN to much larger out-of-distribution graphs. {\color{rev}For each PGM in these two datasets, we generate the graph structure and eigenvalues of the GGM independently and then find a covariance matrix that possesses these eigenvalues and complies with the graph structure, as described below.}

{\color{rev}For \textit{GGM-tree}, we construct a random tree structure by converting from a random Pr{\"u}fer sequence \cite{prufer1918neuer}, which is sampled uniformly from the symmetry group $S_{n-2}$ where $n$ is the number of nodes in the tree.
}

For \textit{GGM-all}, we use the random graph generator algorithm WS-flex proposed in \cite{You.Xie.2020} to generate diverse graph structures parametrized by their average shortest path length and cluster coefficient. 
To generate a random graph, we first sample the average degree $k$ and rewiring probability $p$ parameters for the WS-flex algorithm from a uniform distribution over $[2,n-1]$ and $[0,1]$ where $n$ is the number of variables, and then construct a random graph using these parameters.

{\color{rev}We now describe how we generate a precision matrix that complies with any connected graph structure, whether it is a tree or loopy graph.
Starting from a positive-definite diagonal matrix whose eigenvalues are uniformly drawn from $[0.1,10.0]$,} we first apply a random orthogonal rotation to make it dense. Then we iteratively apply Jacobi rotations to zero out elements according to the connectivity matrix until we achieve the desired structure. We focus on inferring the node variances and give all GGMs a mean of zero because BP is known to give exact means in general GGMs, but only makes errors in marginal variances \cite{weiss1999correctness}. 

For a single GGM, the inverse variance of each variable's marginal Gaussian distribution is used as target for supervised learning. 10000 random graphs with 10 variables are generated as a training dataset. 
Additionally, we generate various testing datasets with 10 to 50 variables, each consisting of 2000 random graphs.

\subsubsection{Binary Third-order Factor Graph}\label{dataset:3spin}
The RF-GNN works with general factor graphs, especially those with higher-order interactions, but GGMs have only pairwise interactions and can also be modeled by regular GNNs \cite{li2022learning}. To test our model on graphs with higher-order factors, we construct a dataset composed of binary graphical models with only third-order interactions.
For the sake of exact solutions of marginal probabilities, we choose small binary spin glass models whose structures and connectivity strengths are randomly sampled so that exact marginal probabilities can be calculated by brute-force enumeration. The joint probability mass function is expressed as
\begin{align}
    p(\{\vs\})\! \propto \!\exp\!\Big( \beta \big(\sum_{p=1}^{3}\sum_{i\in \mathds{1}_{\mathcal{V}}} \!\!b_{i,p} s_i^p +\!\!\!\!\!\!\!\!\!\! \sum_{(ijk) \in \mathcal{N}(\mathcal{F})}\!\!\!\!\!\!\!\!\!\!J_{ijk}s_is_js_k\big)\Big).
\end{align}
where $\mathcal{N}(\mathcal{F})$ is the set of neighbor variable indices for each factor. We generate these random factor graphs using a generalization of the WS-flex generator, parameterized by average variable node degree $k_3$ and rewiring probability $p$. Details of the algorithm could be found in the supplementary material. The singleton potential coefficients $b_i^p$ and 3-way coupling coefficients $J_{ijk}$ are randomly drawn from a standard Gaussian distribution. The inverse temperature, $\beta$, is set to $0.5$.

\subsubsection{Continuous Third-order Factor Graphs}\label{dataset:3continuous}

\begin{figure}
    \centering
    \includegraphics[width=\linewidth]{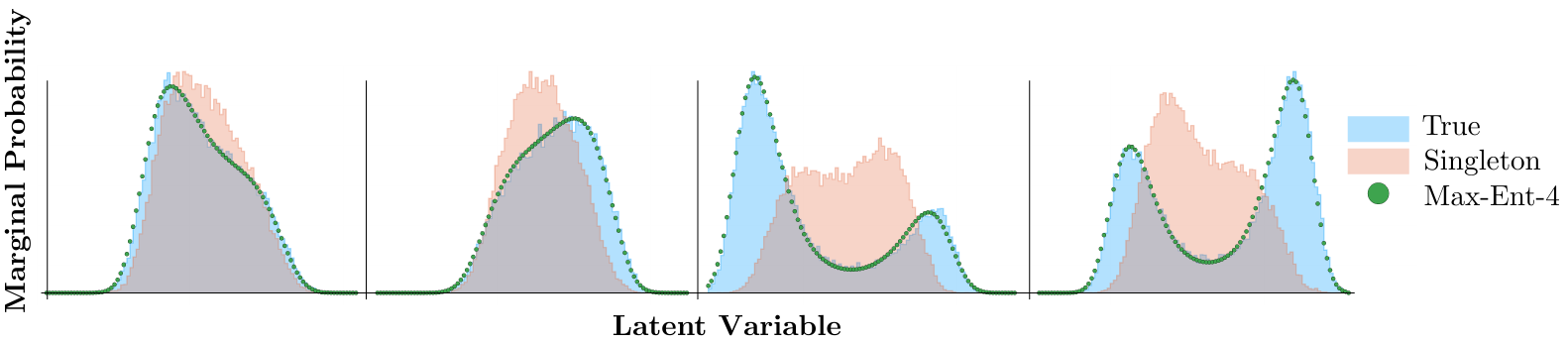}
    \caption{Example univariate marginal distributions from the continuous third-order dataset. Shaded blue curves represents empirical marginal distributions from samples. Shaded orange curves are empirical single-variate distributions by only considering singleton potentials. Green dots represent the maximum entropy fit of the sample marginal distributions by matching the first four central moments}
    \label{fig:marginal}
\end{figure}

We are interested in continuous graphical models with third-order interactions, and we propose to use our method as an approximate inference algorithm for this model class. We construct a dataset of random continuous PGMs with third-order interactions and evaluate the accuracy of our message-passing algorithm against an expensive sampling approach. For each inference problem, a factor graph is randomly constructed such that 3-factors are generated using the aforementioned WS-flex variant, and 2-factors are generated using the regular WS-flex algorithm. Each graph structure is parametrized by three numbers: the average outdegree of variable nodes to 2-factor nodes $k_2$, the average outdegree of variable nodes to 3-factor nodes $k_3$, and the rewiring probability $p$. $k_2$ is uniformly sampled from $[2,n-1]$, $k_3$ is uniformly sampled from $[2,\floor{(n-1)(n-2)/6}]$, and $p$ is uniformly sampled from $[0,1]$. An isotropic 4th-order base measure is added to ensure the joint density is normalizable:
\vspace{-2mm}
\begin{align}
    p(\vx) \propto \exp\Big[ - \beta \Big(\sum_{p=1}^{3}\sum_{i\in \mathds{1}_{\mathcal{V}}} b_{i,p} x_i^p + \!\!\!\!\!\!\!\!\!\sum_{(ij)\in \mathcal{N}(\mathcal{F}_2)}\!\!\!\!\!\!\!\!K_{ij}x_ix_j + \!\!\!\!\!\!\!\!\!\sum_{(ijk)\in \mathcal{N}(\mathcal{F}_3)}\!\!\!\!\!\!\!\!\!J_{ijk}x_ix_jx_k + \lVert \vx \rVert_{\ell_4}^4 \Big) \Big]
\end{align}
The bias parameters $\{b_{i,p}\}$ and interaction strengths $\{K_{ij}\},\{J_{ijk}\}$ are sampled from a standard normal distribution $\mathcal{N}(0,1)$. The inverse temperature $\beta$ is chosen to be $0.3$. 
Note that this exponential family of distributions is not closed under marginalization: marginals are not in the same family as the joint.

As an approximate ground truth for our training algorithms, we run an MCMC algorithm, the No-U-Turn Sampler (NUTS) \cite{hoffman2014no}, using the Stan \cite{standev2021stancore} software for a large number of steps. Our readout targets are summary statistics computed from those generated samples. For each random structure generated, we run 8 MCMC chains with 10000 warmup steps and 10000 sampling steps each for a random set of parameter values, and keep drawing new parameters until the potential scale reduction factor (PSRF) \cite{gelman1992inference} falls below $1.2$, indicating convergence has been reached for the MCMC chains. We target the first four central moments, since the Jensen-Shannon divergence between the empirical sample distribution and corresponding moment-matched maximum-entropy distribution does not decrease substantially when including more moments. \Cref{fig:marginal} shows random examples of sampled univariate marginals, the corresponding singleton distributions obtained by ignoring all multivariate interactions, and their maximum entropy counterparts by matching the first four central moments. Observe that there are substantial differences between these marginals and their singleton potentials. These differences are caused by the influences of other nodes on the graph, and we would like our RF-GNNs to capture these network effects.

\subsection{Training}\label{sec:training}
We implement the RF-GNN\footnote{Code is available at \url{\codelink}} in PyTorch \cite{paszke2019pytorch} and PyTorch Geometric \cite{Fey/Lenssen/2019} and perform all experiments on internal clusters equipped with NVIDIA GTX 1080Ti and Titan RTX GPUs. We randomly split each dataset into a training and a validation set of ratio 4:1. The testing sets are constructed separately for different graph sizes. In every experiment, we use the ADAM \cite{Kingma.Ba.2014} optimizer with batch size 64 and initial learning rate $0.001$. We multiply the learning rate by a factor of $0.2$ if there is no improvement in 20 epochs, and perform early-stopping if there is no improvement in 40 epochs. 
We choose the dimension of hidden states for both variable and factor nodes $\vh_v,\vh_f$ to be 64. 
{\color{rev}All MLP modules use SELU activations \cite{klambauer2017self} and 2 hidden layers with 64 units per layer.}
We use mean-squared error (MSE) as loss on single-variate marginal precision values for the Gaussian dataset and on the first four central moments calculated from MCMC samples for the continuous third-order factor graph dataset respectively. For the binary third-order dataset, we use binary cross entropy as loss.

\section{Experiments and Results}\label{sec:experiment}
All RF-GNNs are trained on graphs of size 10 only, but are tested on graphs of various sizes. We uniformly draw the number of recurrent steps $T$ from $[30,50]$ during training. $T$ is set to 30 during testing. We repeat all experiments 10 times from different random seeds, pick the best model with the lowest validation error and report the bootstrapped mean and 95\% confidence interval on held-out test sets with different graph sizes.
{\color{rev}As a baseline to show how strong the interactions are in each dataset, we also train a separate RF-GNN (singleton RF-GNN) that tries to predict the same target, but only sees a modified graphical model with all interaction terms removed}. A singleton RF-GNN would have lower performance on graphs whose marginal distributions are strongly affected by multivariate interactions.
We show in-distribution results of BP, the full RF-GNN, and the singleton RF-GNN in \Cref{tab:result-mse} and \Cref{tab:result-r2}. {\color{rev}To illustrate the absolute and relative errors, we report Mean Squared Error (MSE) and Coefficient of determination ($R^2$) separately in \Cref{tab:result-mse} and \Cref{tab:result-r2}.}

\rowcolors{1}{white}{white}
\begin{table}[h]
\begin{center}
\caption{Experimental results (Mean Squared Error, MSE) on four synthetic PGM datasets. MSE is used as the training objective for datasets \textit{GGM-tree}, \textit{GGM-all}, and \textit{3-continuous}, while binary cross-entropy loss is used for \textit{3-binary}. Numbers in the parenthesis indicates the uncertainty of the last decimal.}\label{tab:result-mse}
\begin{tabular*}{\textwidth}{@{\extracolsep{\fill}}lSSSS@{\extracolsep{\fill}}}
\toprule%
Method & \text{GGM-tree} & \text{GGM-all} & \text{3-binary} & \text{3-continuous} \\
& \scalelabel{-3} & \scalelabel{-3} & \scalelabel{-3} & \scalelabel{-1} \\
\midrule
BP & 0 & 141.6 & 9.4 & {\tiny N/A}\\
RF-GNN & 4.4 \pm 1.3 & 4.6 \pm 0.1 & 5.10 \pm 0.01 &  1.329\pm 0.006  \\
Singleton & {\tiny N/A} & 902.7 \pm 0.6  & 21.5\pm 0.0 &  1.930\pm0.0\\
\bottomrule
\end{tabular*}
\end{center}
\end{table}

\begin{table}[h]
\begin{center}
\caption{Experimental results (Coefficient of determination $R^2$) on four synthetic PGM datasets. Numbers in the bracket indicates the uncertainty of the last decimal.}\label{tab:result-r2}
\begin{tabular*}{\textwidth}{@{\extracolsep{\fill}}lSSSS@{\extracolsep{\fill}}}
\toprule%
Method & \text{GGM-tree} & \text{GGM-all} & \text{3-binary} & \text{3-continuous} \\

\midrule
BP  & 1 & 0.9719 & 0.8290  & {\tiny N/A}\\
RF-GNN  & 0.9995 \pm 0.0001  & 0.9991\pm 0.0000 & 0.9378\pm 0.0007 & 0.8158\pm0.0008\\
Singleton  & {\tiny N/A} & 0.8329 \pm 0.0000 & 0.7352\pm 0.0000 & 0.6543\pm0.0001 \\
\bottomrule
\end{tabular*}
\end{center}
\end{table}

\begin{figure*}[h]
    \centering
    \makebox[\textwidth][c]{\includegraphics[width=1.\textwidth]{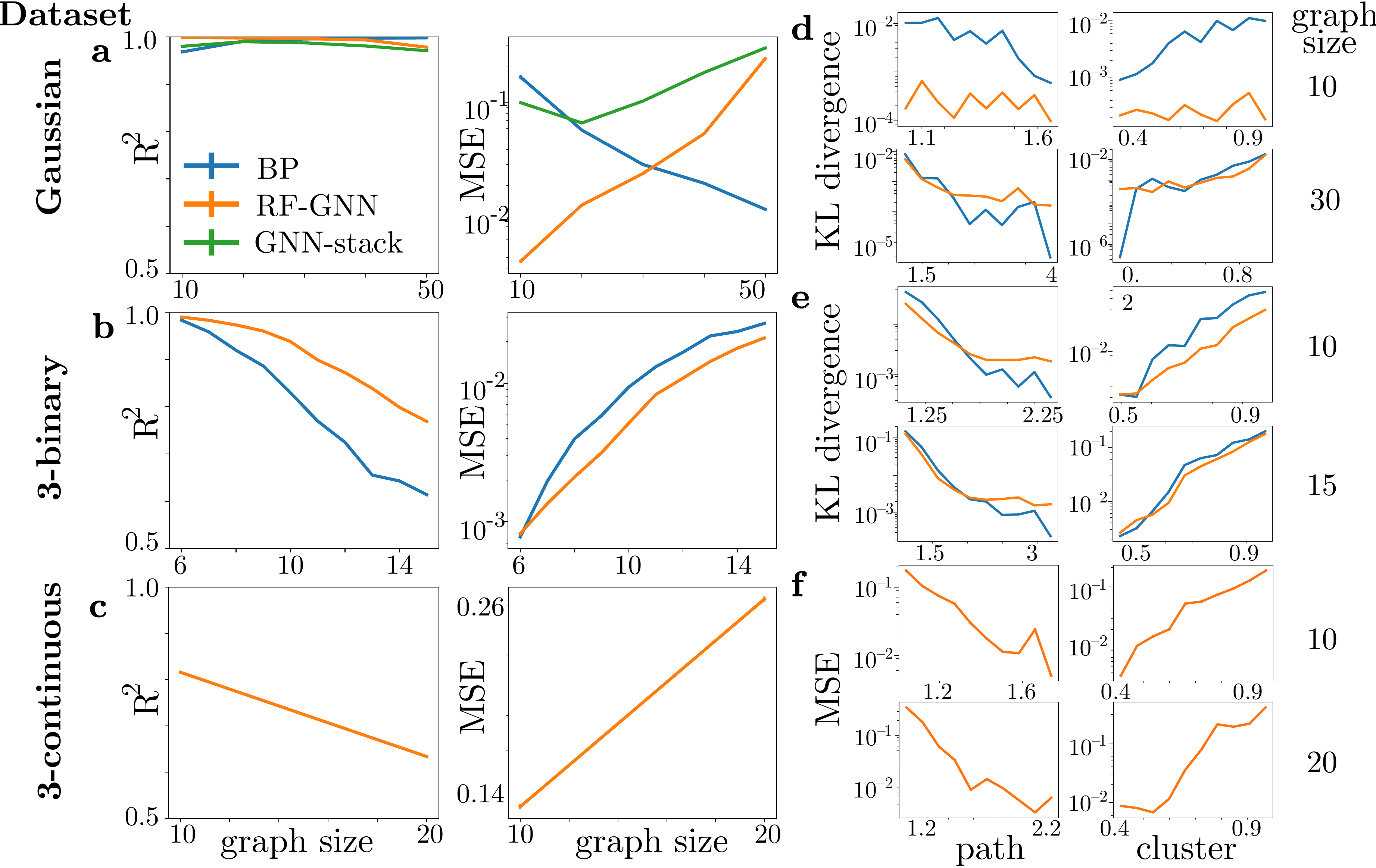}}
    \caption{Performance on Gaussian, third-order spin, and third-order continuous datasets. Row a+d, b+e, and c+f show the results of Gaussian, binary spin, and third-order continuous dataset respectively. In subfigures a,b, and c, we compare the generalization performance of the RF-GNN and BP to larger graphs. All RF-GNNs are trained on graphs with size 10. We also consider an additional 10-layer stacked GNN for the Gaussian dataset as a comparison. Two metrics---$R^2$ and MSE---are shown for the Gaussian and spin datasets. Subfigures d,e, and f show how the error of different methods depends on graph structures. We plot error against the average shortest path length in the left panels, and against the cluster coefficient in the right panels. For each dataset, we show the in-sample results in the top row and generalization results on larger graphs in the bottom row}
    \label{fig:ggm-spin}
\end{figure*}

We also show in \Cref{fig:ggm-spin} the performance of RF-GNNs when generalizing to unseen graph structures with different sizes from the same parametric family. 
{\color{rev}It is well known that Belief Propagation becomes non-exact on loopy graphs. Here, we empirically investigate how BP and an RF-GNN performance depends on the graph structure by examining graph metrics.  
We choose two independent structural features that quantify the loopiness of a graph: average shortest path length and cluster coefficient \cite{watts1998collective}. 
The former quantifies the average loop length between any two nodes, and the latter describes the small-world property of a graph.
Graphs in each test set are binned into 10 equal bins according the graph metric, and for each bin we report the bootstrapped mean and its 95\% CI for BP and RF-GNN performance metrics in (\Cref{fig:ggm-spin}\textit{d-f}). Many real-world graphs have a small average path length and a large cluster coefficient \cite{watts1998collective}, and this is the region where our model outperforms BP. 
}

\subsection{Gaussian Graphical Models}\label{results:ggm}

\subsubsection{Tree Graphs}\label{results:ggm-tree}
{\color{rev}We trained a model on a dataset of 10000 Gaussian tree graphs with size 10 using the eigenvalue distribution in \Cref{dataset:ggm}. 
BP is exact on trees, but our model also achieves excellent generalization performance even when generalizing to graphs with size 50, giving an in-distribution $R^2$ score $0.9995\pm 0.0001$ and out-of-sample $R^2$ score $0.9996\pm 0.0002$.}

\subsubsection{Loopy Graphs}\label{results:ggm-all}
In this experiment we tested the performance of our model on Gaussian Graphical Models with zero mean, random precision matrices, and various structures as described in \Cref{dataset:ggm}. 
\Cref{fig:ggm-spin}\textit{a}, right, shows that our model trained on a dataset of GGMs with 10 variables has an average MSE that is 30-fold smaller than Belief Propagation on a test set with the same size. 
When generalizing to larger graphs, our model still has a smaller error than BP on graphs up to 30 variables. We also report the corresponding $R^2$ score as a metric for goodness of prediction in \Cref{fig:ggm-spin}\textit{a}, left. 
Since BP doesn't always converge on loopy graphs, all metrics are calculated conservatively using the subset of test graphs with convergent BP dynamics, which comprise from 73.2\% to 71.2\% of the test set as the graph size varies from 10 to 50. We consider BP to be non-convergent if the absolute error of beliefs between two adjacent BP updates still exceeds $10^{-5}$ after 1000 cycles.


BP can be viewed as a recursive algorithm that finds local fixed points of the Bethe free energy upon convergence \cite{heskes2003stable}. That is why we also construct our method as an autonomous dynamical system. However, for comparison, we also constructed a feedforward model with 10 Factor-GNN layers stacked together. \Cref{fig:ggm-spin}\textit{a} shows that although the feedforward model has almost 10 times more parameters, the performance is worse than the recurrent on test datasets with graph size varying from 10 to 50.

\subsection{Binary Third-order Graphical Model}\label{result:3spin}

For the binary graphs with only third-order interactions and no pairwise interactions, we train an RF-GNN on a dataset of size 10 and test on graphs with sizes from 6 to 15. We didn't test on larger graphs because it becomes impractical to enumerate all spin configurations in order to compute exact marginal probabilities. 
For binary graphical models, unlike in \Cref{results:ggm}, non-convergent BP dynamics won't diverge, but may oscillate. Thus, we take the beliefs from the last cycle if the dynamics does not converge in 1000 cycles. There is no qualitative difference when evaluating model performance using the whole test dataset or just BP-convergent ones, so we report the metrics on BP-convergent graphs to maintain consistency with \Cref{results:ggm}.
Within the range of testing graphs, RF-GNNs consistently outperform BP on average (\Cref{fig:ggm-spin}\textit{b}). 
In the space of graph structures, RF-GNNs perform better than BP in regions with smaller average shortest path length and larger cluster coefficient, which agrees with the result of \Cref{results:ggm} (\Cref{fig:ggm-spin}\textit{e}). 

\subsection{Continuous Third-order Graphical Model}\label{result:3cont}
For general, continuous, non-Gaussian graphical models it is not feasible to compare the RF-GNN with BP because the BP message update is neither an explicit formula, nor calculable exactly by enumeration. Even EP message updates need to be approximated \cite{Heess.Winn.2013,Eslami.Winn.2014,Jitkrittum.Szabo.2015}. 
These methods approximate local BP updates without knowing the global graph structure, thus inheriting the drawbacks of BP on loopy graphs. Instead of approximating the EP message-passing for a specific type of factors within the EP framework, we learn a new message-passing algorithm end-to-end that works with many factor types and avoids some of EP's limitations. This approach is more applicable to loopy graphs where BP and EP struggle.

For this experiment, we constructed a continuous graphical model with third-order interactions and train our model to predict the first four central moments of every univariate marginal distribution (see \Cref{dataset:3continuous}).
Since BP or EP is not feasible for this dataset, we only compare the RF-GNN with the singleton RF-GNN as our baseline.
An RF-GNN trained on graphs with 10 variables achieves an in-sample $R^2$ score of $0.816\pm 0.001$, while the baseline model without interactions has an $R^2$ of only $0.654\pm 0.001$. We test the generalization performance of our model with a dataset of graphs of size 20. For the MCMC algorithm to produce convergent chains for these larger graphs, we shrink the interaction strengths by a factor of 2 while keeping the singleton parameter distribution unchanged so that both input and output values fall in the training ranges. On a test set of 1000 graphs,
the full model has an $R^2$ score of $0.654\pm 0.000$, while the baseline model has an $R^2$ score of $0.495\pm 0.000$. Qualitatively similar to the Gaussian and spin datasets, model performance drops when generalizing to larger graphs (\Cref{fig:ggm-spin}\textit{c}), and our model makes larger errors on loopy graphs (\Cref{fig:ggm-spin}\textit{f}) since these are harder problems.

\subsection{Low-density Parity-check (LDPC) Codes}\label{result:ldpc}
As one simple real-world application of our method, we tested the RF-GNN on a Low-Density Parity-Check (LDPC) decoding dataset used for the Factor Graph Neural Network (FGNN) \cite{Zhang.Lee.2019eqj} and compare the performance of an RF-GNN and FGNN along with other baseline models used in \cite{Zhang.Lee.2019eqj}.
{\color{rev}In the LDPC dataset, a noisy signal $\tilde{\mathbf{y}}$ is obtained by transmitting the original signal $\mathbf{y}$ through a noisy channel with Gaussian noise and irregular bursting noise:}
\begin{align}
    & \tilde{y}_i = y_i + n_i + p_iz_i \\ 
    & n_i \sim \mathcal{N}(0,\sigma^2) \\ 
    & z_i \sim \mathcal{N}(0,\sigma_b^2) \\ 
    & p_i \sim \text{Bernoulli}(0.05)
\end{align}
The task is to decode the first 48-bit or original signal from a 96-bit noisy signal $\tilde{\mathbf{y}}$. The graphical model is fixed in this task given a specific coding scheme, with 96 variables each connected to three factors, and 48 parity-check factors each connected to six variables. 

\begin{figure*}[htbp]
    \centering
    \includegraphics[width=\textwidth]{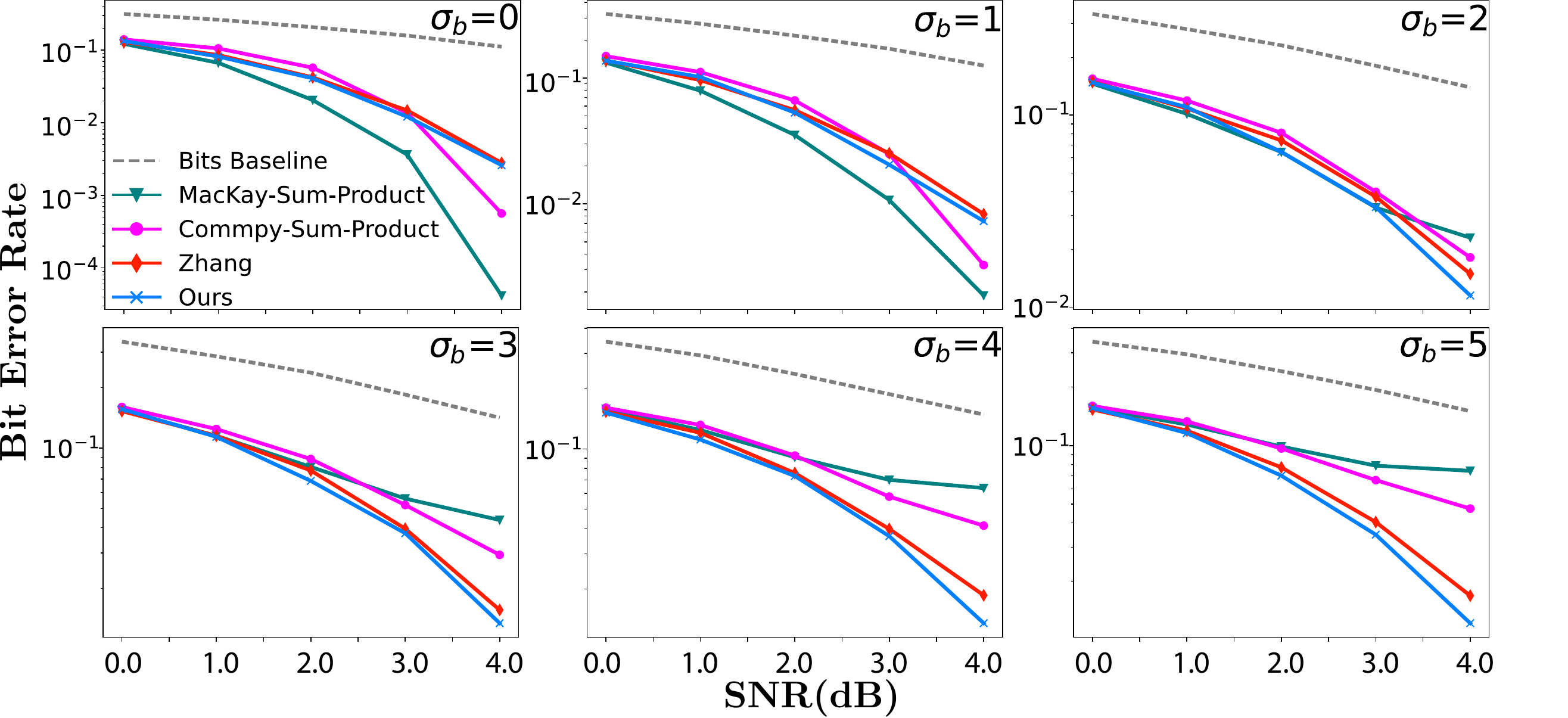}
    \caption{Performance curves of several models \cite{mackay2009david,taranalli2015commpy,Zhang.Lee.2019eqj} on the LDPC Decoding Dataset, showing the bit error rate as a function of the independent Gaussian noise $\sigma$ for different values of burst noise $\sigma_b$. SNR in decibels is calculated as $-10\log_{10}(\sigma^2)$. All shown methods outperform the na\"{\i}ve baseline of decoding bits independently (dashed line), but our method exhibits the best improvement when the burst noise is high}
    \label{fig:ldpc}
\end{figure*}

We trained an RF-GNN using the hyperparameters stated in \Cref{sec:training}, except in this specific task we use a fixed number of 30 iterations instead of a random number for better performance, since generalization to different graph structures is not needed in this task. The performance curves of the RF-GNN and compared models for different levels of burst noise are presented in \cref{fig:ldpc}. Testing results for competing methods are directly taken from \cite{Zhang.Lee.2019eqj}'s public figure generation scripts on GitHub. The testing dataset is also generated with \cite{Zhang.Lee.2019eqj}'s code with fixed random seeds. Due to the recurrent nature of our method, it currently takes more clock time to process one batch compared to the feedforward FGNN. However, our model converges with 100$\times$ fewer iterations, after seeing only $10^6$ samples, while the FGNN converges after seeing $10^8$ samples. While being more parameter-efficient, our model has similar error rates as FGNN under low SNR and smaller error rates at high SNR.

\section{Related Work}
Previous work has explored learning to pass messages that calculate marginal probabilities in graphical models. Some examined fast and approximate calculation of Belief Propagation or Expectation Propagation updates when analytical integration is not feasible for computing on the level of single factor \cite{Heess.Winn.2013, Eslami.Winn.2014, Jitkrittum.Szabo.2015}. Others created message-passing algorithms for inference in probabilistic graphical models, trying to learn an algorithm that is more accurate than Belief Propagation when the underlying graph is loopy \cite{Yoon.Pitkow.2018}.
The most related work to ours uses stacked bidirectional GNNs on factor graphs (FGNNs) to perform maximum {\em a posteriori} estimation on binary graphical models \cite{Zhang.Lee.2019eqj}. One major difference is that we use a recurrent network to construct a dynamical system that runs to convergence, like Belief Propagation, instead of a feedforward network with fixed number of layers, so our method uses far fewer parameters and scales to larger graphs without adding new layers. Another related study puts a factor graph NN layer within a recurrent algorithm to calculate marginal probabilities in binary graphical models, applying them to low density parity check codes and Ising models \cite{Satorras.Welling.2020}. Instead of using a GNN to learn a novel message-passing algorithm from scratch, the authors build their algorithm on top of Belief Propagation and let the GNN serve as an error-correction module for loopy BP.

Some work has been done to extend GNNs to such higher-order graphs \cite{Bai.Torr.2019,Zhang.Ye.2019}, including studies aiming to solve graph isomorphism tasks \cite{Morris.Grohe.2019}, and others applying GNNs to other high-order data structures like simplicial complexes \cite{Ebli.Spreemann.2020}.

\section{Discussion}\label{sec:discussion}
\subsection{Limitations}

{\color{rev}A limitation of our method is that a separate RF-GNN needs to be learned from scratch whenever encountering a new parametric family of graphical models. One exciting but speculative possibility is to extend this framework such that whenever a new type of factor is encountered, the model only trains a dedicated new encoder/decoder, and reuses a shared core message-passing module, perhaps implementing a more universal inference engine akin to a shared language model \cite{denkowski2014meteor}.}
This would be useful when data is scarce, since the difficult message-passing core could be learned from multiple rich datasets. The general formulation of Belief Propagation takes such a universal form, as have some multi-factor extensions like Generalized BP \cite{yedidia2000generalized}. We hypothesize that the RF-GNN could potentially extend this universality while compensating for some challenges of inference in a loopy world. 

As discussed and explored by \cite{chen2019equivalence, sato2020survey,maron2019provably,keriven2019universal} and others, message-passing algorithms with only local information like BP and regular GNNs, including the RF-GNN, would fail to solve some graph isomorphism tasks on loopy graphs. These algorithms will produce the same marginals for two locally similar but globally different PGMs, even though the true marginals should be different. Some GNN variants \cite{murphy2019relational,Morris.Grohe.2019,chen2019equivalence} allow nodes to see beyond their direct neighbors in order to partially mitigate this issue. We hypothesize these could be employed as a replacement of local GNNs to get better inference performance on loopy graphs, at the costs of graph locality and greater complexity.

\subsection{Extensions}
In this paper, we show that our recurrent factor graph neural network is a valuable alternative to traditional message-passing algorithms like Belief Propagation as an approximate inference engine on graphical models. It provides greater flexibility that accommodates inference on a far richer class of graphical models with multi-way interactions.
It outperforms BP on some loopy graphs, and offers a practical way to perform fast inference for continuous non-Gaussian graphical models.

Although our method achieves promising results and does generalize to graphs of different sizes, it naturally performs better within its training distribution. However, we saw almost no performance drop on Gaussian tree graphs when generalizing to 5$\times$ larger graphs (\Cref{results:ggm-tree}), and we saw promising generalization results on loopy graphs (\Cref{results:ggm-all}, \Cref{result:3spin}). This suggests that our methods could be used for inference on large graphs by training on smaller ones.
{\color{rev}Although our method needs training data, general MCMC sampling methods like NUTS \cite{hoffman2014no} could always be used to generate training data on small graphs when efficient sampling methods for larger ones are not available.}
Compared to a 10-layer stacked GNN (\Cref{fig:ggm-spin}\textit{a}), our method achieves better performance with substantially fewer parameters. However, one limitation is that in our experiment we simply stack the Factor-GNN layers without any further engineering tricks like skip connections, pooling layers, etc., unlike \cite{Zhang.Lee.2019eqj}. Our recursive message-passing algorithm has a substantial advantage over a feedforward one when generalizing to larger graphs: the number of layers must scale with graph size for a feedforward network in order to distribute information throughout the graph, whereas a recurrent algorithm can run until convergence regardless of the graph size. 

{\color{rev}Besides performing marginalization, RF-GNNs could also be trained to calculate the most probable state vector by altering the training objective to treat simultaneous local readout as a global target.
Conditioning should also be straightforward since by conditioning on some subset of variables, we get a smaller PGM within the same family. Thus, the same RF-GNN could be directly applied to perform inference on the conditional model.
}

While belief propagation is exact on trees, it is only an approximation when used on loopy graphs because the update rule incorrectly assumes that messages are independent, as they are for tree graphs. The topological structure of the graph therefore has an important impact on the performance of these algorithms. Past work has attempted to mitigate this effect, creating new message-passing algorithms by training GNNs to compensate \cite{Yoon.Pitkow.2018}. Note that it is not only the existence of loops that creates problems, but also the length of the loops and the strength of interactions along those loops. Since topology is insensitive to the size of loops, it would be blind to these effects, but topological data analyses often look for persistent homology, which is sensitive to the sizes of loops. Often, it is shorter loops that create the most problems in belief propagation. Thus, by incorporating higher-order structure, one may absorb smaller loops into cliques, potentially improving performance while preserving the large-scale topology.

{\color{rev}
Higher-order cliques are an important ingredient in other computations as well. One notable example is the simplicial complex, a foundational object in algebraic topology. Simplicial complexes are collections of fully-connected cliques that recursively contain all fully-connected subcliques among all subsets of nodes. The structure of simplicial complexes can reveal large-scale topological features of a graph, and GNNs can be applied to simplicial complexes \cite{bodnar2021weisfeiler1,bodnar2021weisfeiler2,ebli2020simplicial}. However, using simplicial complexes as a primitive requires us to pass more messages to account for dense constraints across all neighboring orders: 4-cliques must interact with all contained 3-cliques, which interact with all contained 2-cliques, and so on. In contrast, there may be major computational advantages to using sparser messages between subsets of cliques. This could be because there are indeed sparse higher-order interactions in the graphical model (as the PGMs in our study) without the inclusion structure implied by simplicial complexes. Alternatively, sparse higher-order messages can be helpful even when the true model has low-order interactions because it can reduce the number of messages as seen in Kikuchi clustering \cite{yedidia2000generalized}, Junction Tree \cite{lauritzen1988local}, and higher-order Weisfeiler-Leman algorithms \cite{Morris.Grohe.2019}.}

{\color{rev}In this paper, we focus on performing inference on known graphical models. However, in many real-world applications, it is also hard to estimate the model parameters from data. One way an approximate message-passing inference algorithm like ours could facilitate parameter estimation is through self-supervised training like in \cite{lazaro2021query}. Given some data that is assumed to be generated by a graphical model we are fitting, we could construct a new graphical model by randomly masking a portion of observed data as hidden and use GNN on the currently fitted graphical model to infer the masked-out data. In this way, both the model parameters and the GNN could be trained to make high quality partial inferences. However, the model parameters could be biased towards partial inference instead of Maximum Likelihood Estimation, as shown in \cite{lazaro2021query}.}

\section{Statements and Declarations}

XP is a co-founder of Upload AI, LLC, a company which may have a competing interest in developing related algorithms. YF has no competing interests to declare.

\section{Acknowledgments}
The authors thank KiJung Yoon for helpful conversations about graph metrics and Alex Kunin for conversations about topological spaces. YF and XP were supported in part by NSF NeuroNex grant 1707400. XP was supported in part by NSF CAREER award 1552868 and a grant from the McNair Foundation.



\clearpage
\begin{center}
\textbf{\large Supplementary Materials: Generalization of graph network inferences in higher-order probabilistic graphical models}
\end{center}
\setcounter{equation}{0}
\setcounter{section}{0}
\setcounter{figure}{0}
\setcounter{table}{0}
\setcounter{page}{1}
\makeatletter
\renewcommand{\theequation}{S\arabic{equation}}
\renewcommand{\thefigure}{S\arabic{figure}}
\renewcommand{\thetable}{S\arabic{table}}
\renewcommand{\thesection}{S\arabic{section}}

\section{Random factor graph generator for 3-factors}
We use a variant of the WS-flex algorithm \cite{You.Xie.2020} to generate a random factor graph with only 3-factors. This includes two stages: generating a ring-like factor graph, and rewiring. In the first stage, given average variable node degree $k$ and number of variable nodes $n$, the number of factors is $f=\lfloor n  k / 3\rfloor$. On average, there are $\floor{f/n}$ 3-factors centered at each variable node. The other two legs of these factors are connected to the closest neighbors of the center variable, while keeping the factor unique. For example, three 3-factors centered at variable 5 may be connected to variables $(4,5,6)$, $(4,5,7)$ and $(3,5,6)$ respectively. The remaining $(f \text{ mod } n)$ 3-factors are then formed by randomly picking the center variable and choose the closest two neighbors, while avoiding redundant factors. 
After this ring is created, in a second stage we may rewire each factor with probability $p$ by randomly picking three new neighbors without duplication.

\section{Additional results}

As a supplement to the main results presented in \Cref{fig:ggm-spin}, \Cref{tab:result-mse} and \Cref{tab:result-r2}, here we show some additional data and figures which support the results shown in the main text.

\begin{figure*}[htbp]
    \centering
    \includegraphics[width=.5\textwidth]{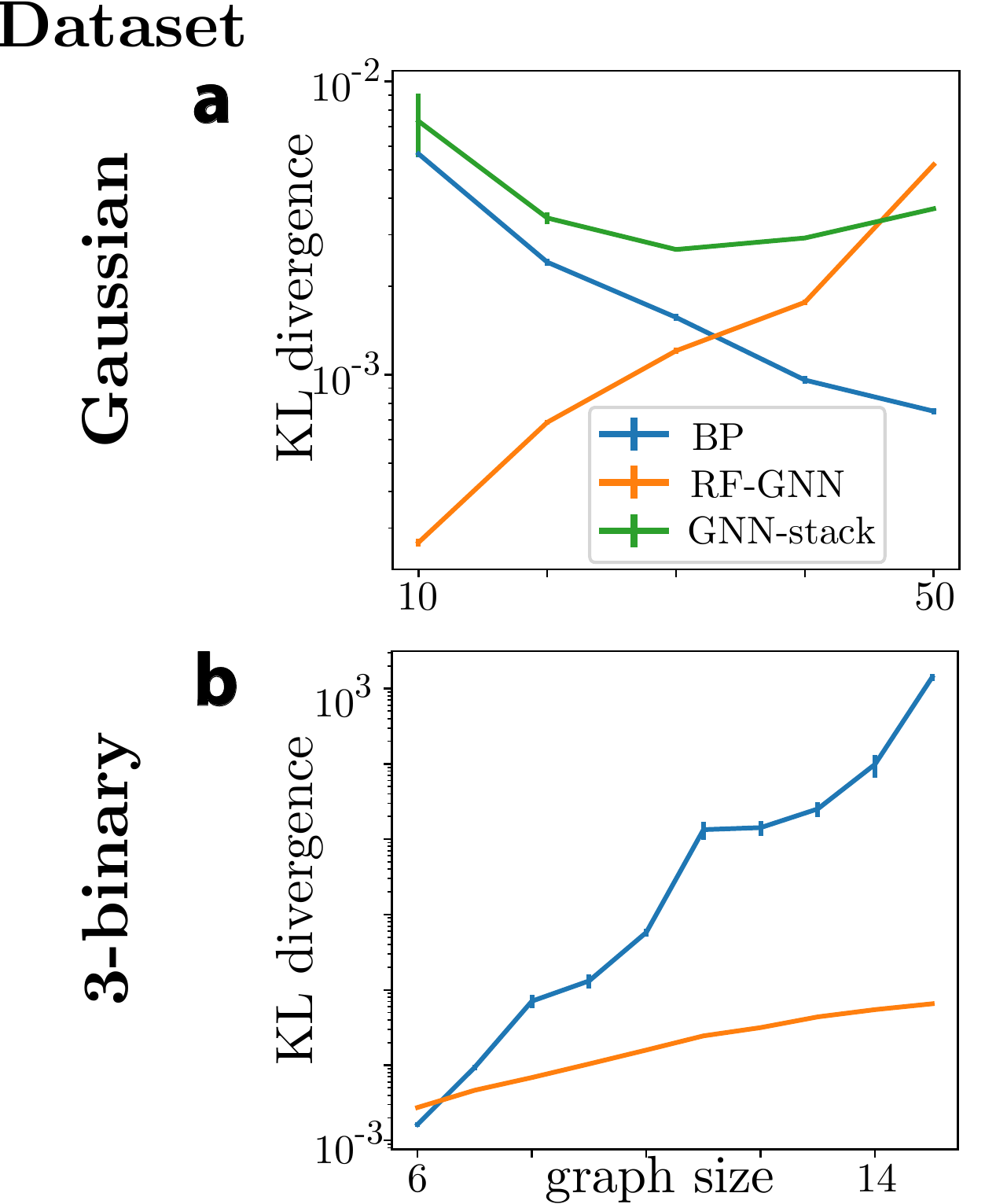}
    \caption{Generalization performance on two PGM datasets for which exact KL-divergence between ground truth and predicted distributions could be calculated.}
    \label{fig:result-supp}
\end{figure*}

In additional to MSE and $R^2$ score, we also compute the KL divergence between the true marginal and the predicted marginal for applicable datasets. Results are shown in \Cref{tab:result-kl} and \Cref{fig:result-supp}, which are in the same formats as \Cref{tab:result-mse} and \Cref{fig:ggm-spin}.

\begin{table}[h]
    \begin{center}
    \begin{minipage}{\textwidth}
    \caption{Experimental results (KL-divergence between true marginal and the predicted distribution) on three applicable synthetic PGM datasets. We omit the KL-divergence calculation for dataset 3-continuous as we don't have exact true marginal distributions}\label{tab:result-kl}
    \begin{tabular*}{\textwidth}{@{\extracolsep{\fill}}lSSSS@{\extracolsep{\fill}}}
    \toprule
    Method & \text{GGM-tree} & \text{GGM-all} & \text{3-binary} & \text{3-continuous} \\
    & \scalelabel{-3} & \scalelabel{-3} & \scalelabel{-3} & \scalelabel{-4} \\
    \midrule
    BP &0 & 6.5 & 26.9 & {\tiny N/A}\\
    RF-GNN & 1.3 \pm 0.4 & 0.27\pm 0.01 & 15.9\pm 0.2 & {\tiny N/A}\\
    Singleton & {\tiny N/A} & 46.0\pm 0.1 & 52.8 \pm 0.0 & {\tiny N/A}\\
    \bottomrule
    \end{tabular*}
    \end{minipage}
    \end{center}
\end{table}

\begin{figure*}[htbp]
    \centering
    \includegraphics[width=.9\textwidth]{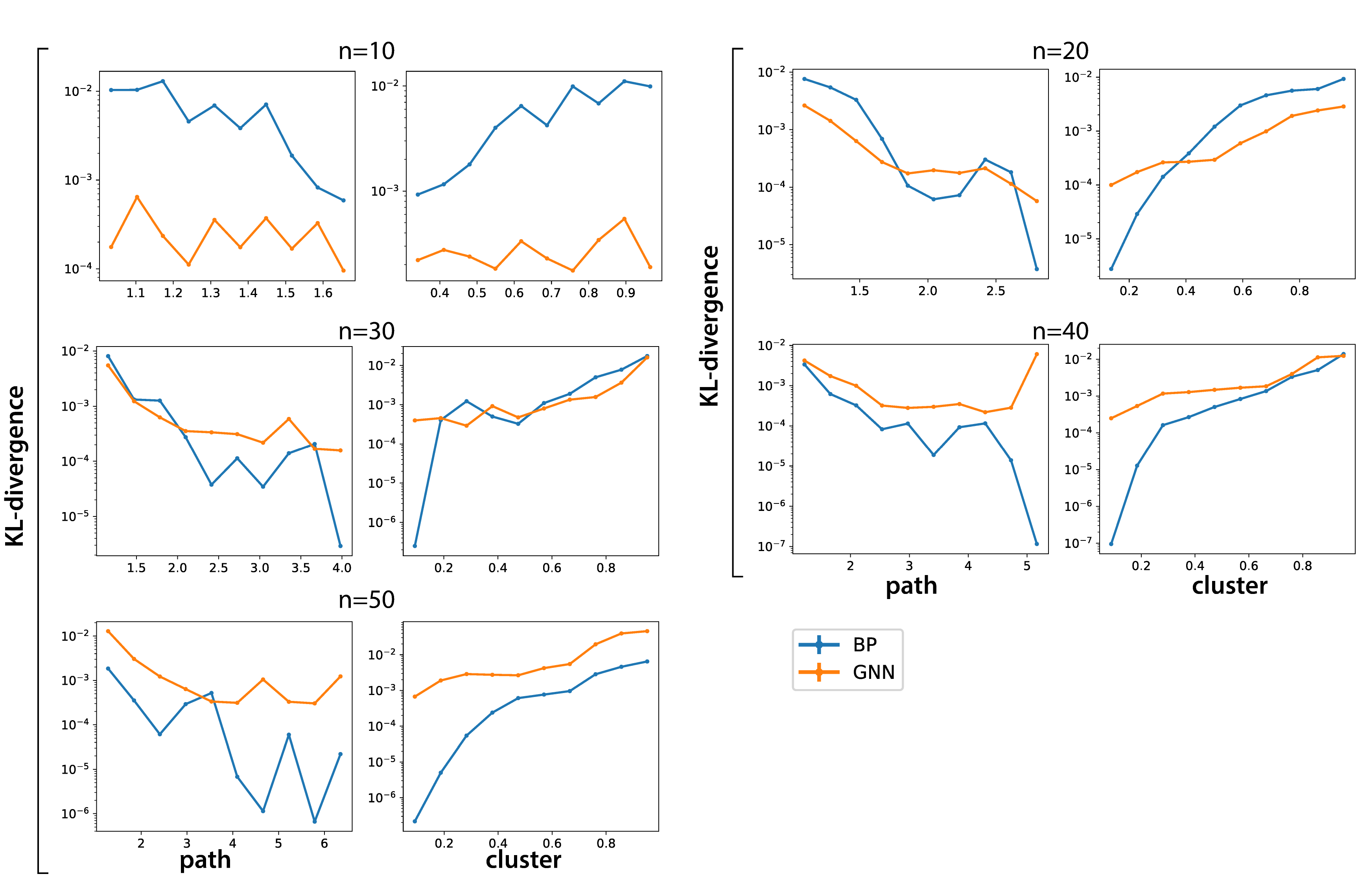}
    \caption{\textbf{Gaussian Dataset:} dependence of KL divergence between true marginal distribution and model prediction on average shortest path length and cluster coefficient. We compare Belief Propagation (blue) and an RF-GNN (orange). The RF-GNN is trained on graphical models with 10 variables. Each pair of panels shows the testing performance on a separate dataset with graphs of fixed size ranging from 10 to 50.}
    \label{fig:gaussian-graphmetric}
\end{figure*}

\begin{figure*}[htbp]
    \centering
    \includegraphics[width=.9\textwidth]{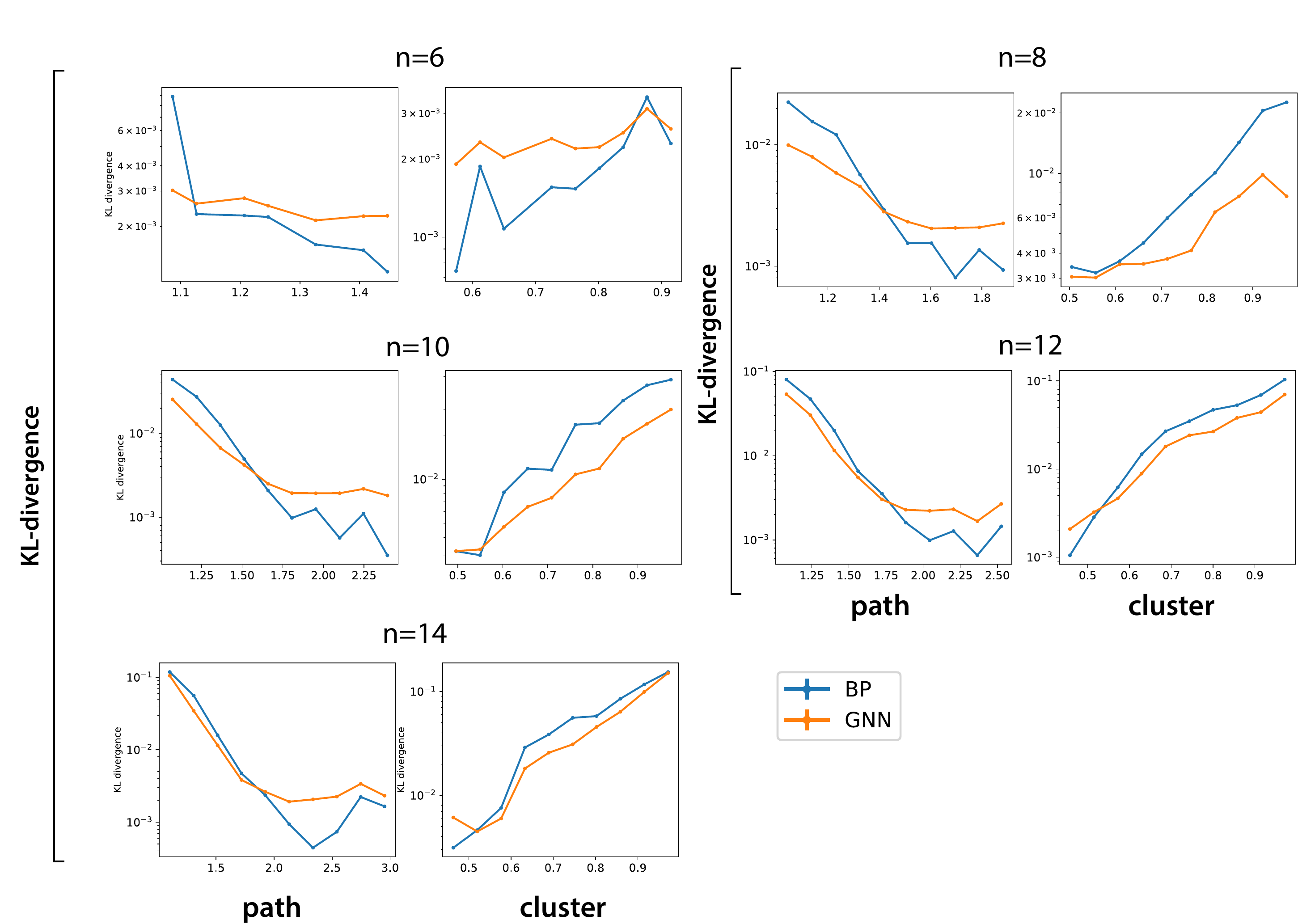}
    \caption{\textbf{Binary Dataset:} dependence of KL divergence between true marginal distribution and model prediction on average shortest path length and cluster coefficient, plotted as in Supplemental Figure \ref{fig:gaussian-graphmetric}. Each pair of panels shows the testing performance on new graphical models with a fixed number of variables ranging from 6 to 14.}
    \label{fig:3spin-graphmetric}
\end{figure*}

In the main text, we show how GNN and BP perform with respect to two graph metrics: average shortest path length and cluster coefficient. Here in \Cref{fig:gaussian-graphmetric} and \Cref{fig:3spin-graphmetric} we provide two figures on testing sets with more sizes for the Gaussian dataset and the binary dataset.

\clearpage
\bibliographystyle{abbrv}
\bibliography{main}
\end{document}